\pdfoutput=1

\documentclass[11pt]{article}

\usepackage[review]{acl}

\usepackage{times}
\usepackage{pgfplots}
\usepackage{fancyhdr} 
\usepackage{latexsym}
\usepackage[normalem]{ulem}
\usepackage[T1]{fontenc}

\usepackage[utf8]{inputenc}

\usepackage{microtype}

\usepackage{caption}
\usepackage{subcaption}
\usepackage{inconsolata}
\usepackage{hyperref}

\newcommand{\thickhline}{\noalign{\hrule height 1pt}}
\title{A Continued Pretrained LLM Approach for Automatic Medical Note Generation}

\author{%
Dong Yuan* \and Eti Rastogi* \and Gautam Naik \and Sree Prasanna Rajagopal
\AND
Sagar Goyal \and Fen Zhao \and Bharath Chintagunta \and Jeff Ward
\\ 
\textit{DeepScribe Inc.}\\
\textit{San Francisco, California, USA}\\
\textit{\{dong, eti, gautam, sree, sagar, fen, jai, jeff\}@deepscribe.tech}\\
}

\begin{document}
\makeatletter
\newif\ifacl@finalcopy
\acl@finalcopytrue
\makeatother

\maketitle

\footnotetext[1]{*Core Contributors and Corresponding Authors}
\begin{abstract}
LLMs are revolutionizing NLP tasks. However, the use of the most advanced LLMs, such as GPT-4, is often prohibitively expensive for most specialized fields. We introduce HEAL, the first continuously trained 13B LLaMA2-based LLM that is purpose-built for medical conversations and measured on automated scribing. Our results demonstrate that HEAL outperforms GPT-4 and PMC-LLaMA in PubMedQA, with an accuracy of 78.4\%. It also achieves parity with GPT-4 in generating medical notes. Remarkably, HEAL surpasses GPT-4 and Med-PaLM 2 in identifying more correct medical concepts and exceeds the performance of human scribes and other comparable models in correctness and completeness. 
\end{abstract}


\section{Introduction}




The emergence of large language model (LLM) has brought revolutionary changes to natural language processing and understanding tasks, paving the way for practical applications of AI across multiple domains such as law, finance, and healthcare. Private LLMs such as GPT-4~\citep{OpenAI2023GPT4TR} and Med-PaLM 2~\citep{singhal2023towards} and open-source LLMs like LLaMA2~\citep{Touvron2023Llama2O} have shown strong performance on general NLP benchmarks. However, recent studies have shown promise that with continued training on more targeted datasets, e.g. smaller LLMs like Orca~\citep{orca, orca2} and Phi-2~\citep{phi2}, 
can surpass much larger LLMs on general tasks.
Despite the success of LLM in general capabilities, they often fall short in niche domains like healthcare, where precision and profound understanding are crucial.
Hence, several models such as Meditron-70B~\citep{chen2023meditron}, PMC-LLaMA~\citep{wu2023pmc} have emerged.



Transcribing medical conversations is a challenging task for both humans and machines due to potential transcription errors and the innate complexity of spoken language, an issue unaddressed by existing medical LLMs.
Existing LLMs trained on medical data largely do well on problems like medical Q\&A but struggle to produce a comprehensive EHR-compatible medical note. Some domain-adapted LLMs \citep{clinical-text-summ} can write some components of the note, but they leave out the crucial "Subjective" section. Some fine-tuned models \citep{leveraging-pretrained} can generate notes from medical conversations but need human overview.

Overall, we developed a new medical LLM proficient in interpreting medical conversation. By using techniques like continued pretraining on diverse data and explanation tuning, including medical and general web corpora, GPT-4 task instructions, EHRs, the model 
was capable of producing medical SOAP notes approved by physicians.

Our main contributions include:

To the best of our knowledge, we are the first to build a small-size (13B) medical LLM that can produce medical notes without any human intervention from doctor-patient conversations that bypass human quality and are accepted by physicians.

HEAL surpasses Med-PaLM 2 and other publicly available models of the same size, matches GPT-4's performance in medical notes generation, and excels with the highest completeness.

Despite having a smaller model size, we achieved an accuracy of 78.4\% on PubMedQA, outperforming GPT-4 and within 5\% of Med-PaLM~2's performance. %

\section{Continued Pretraining}

\subsection{Dataset}

\textcolor{black}{We collected our training data from three major sources} to enable the model to generate coherent English sentences, comprehend medical content, and execute complex instructions required for generating medical notes. (see Table \ref{table:1})


\noindent\textbf{Non-medical public datasets.}
To ensure that the new model doesn't lose the generative capabilities of the pretrained LLaMA2 model, we added general domain datasets such as C4~\citep{c4}. Continued pretraining on them was crucial for generational tasks, enhancing the model's grammar and phrase composition skills.
Initially, we also included filtered subtitle data from open-subtitle 
and youtube. 
However, we decided to exclude these datasets due to their poor quality negatively impacting the model's performance.

\noindent\textbf{Medical public datasets.}
\textcolor{black}{We filtered data from medical web domains such as nih.gov to cover different aspects of medical concept understanding and replay medical knowledge to the model, so the model won't forget the medical knowledge after continued training.} MedDialog~\citep{meddiag} taught medical language conversation while reading materials such as PubMed articles~\citep{pile} provided the model with an overall medical context. 
\textcolor{black}{PubMed and filtered web medical corpus were two major sources, each contributed around 2.5B tokens each in the final training dataset.}


\begin{table}[t]
\begin{center}
\small
\begin{tabular}{c c c}
\thickhline
Dataset & Number of tokens & Percentage of \\
  & (in billions) & total data \\
\hline
Non-medical public & 5.33 & 35.79 \\
Medical public & 5.68 &  38.14 \\
Medical proprietary & 3.88 & 26.07 \\
\textbf{Total} & \textbf{14.89}	& \textbf{100.00} \\
\thickhline
\end{tabular}
\caption {Pretraining datasets.}
\vspace{-15pt}
\label{table:1}
\end{center}
\end{table}

\noindent\textbf{Proprietary medical datasets.}
We also curated a deidentified proprietary medical dataset that consists of real-world doctor-patient conversations from the United States, Electronic Health Records (EHR), SOAP (Subjective, Objective, Assessment, and Plan) notes, 
and ROS (Review of System) templates. 
\textcolor{black}{We also created a synthetic dataset comprising of medical instructions, like extraction of medications from a medical conversation and grammar correction of a generated medical note, respectively. These instructions were generated with the help of both humans and GPT-3.5/GPT-4.} For some of the instructions, we also included {detailed} explanation {as shown in} \cite{orca}. 
\textcolor{black}{Training on such instructions with explanations, helped the model better comprehend the medical notes and understand the reasoning behind it, which was especially needed for the downstream medical documentation task.}
\textcolor{black}{
For example, we created a medical instruction that asks the model to retrieve information from a conversation as shown below:
\begin{quote}
You specialize in summarizing medical conversations, providing clear and thorough explanations so that people can trust your summary with evidence.
I have part of a transcript from a conversation between my doctor and myself. \\
Task: Summarize the \textit{<targeted content>} from this conversation. \\
Requirements: \textit{<requirements>}\\
Transcript: \textit{<transcript>}
\end{quote}
\vspace{-3pt}
Then we further created instructions about reviewing the generated note:
\begin{quote}
Your job is to review a given medical note and generate an updated note. \\
Rules: \textit{<rules on how to review>}. \\
List all the needed updates for the medical note as Updates. Return the updated medical note as Updated Medical Note.\\
Transcript: \textit{<transcript>} \\
Medical Note: \textit{<medical note>}
\end{quote}
\vspace{-3pt}
}
Finally, both of them were used for training the model to improve the model's understanding of the summarization task. 

While we developed a much larger high-quality custom dataset \textcolor{black}{including more than 60B tokens}, currently only 14.89B tokens were used for this training exercise.


\subsection{Training Details}
We performed training using FSDP ~\citep{fsdp} pipeline parallelism with hybrid sharding and flash attention 2 on 32 A100 80 GB GPUs.
We continued training LLaMA2 13B using learning rate of 5e-5 which decays to 1e-5 following a cosine schedule.
We chose a \textcolor{black}{relatively small} batch size of 256, \textcolor{black}{to achieve more than 10K effective gradient update steps.} 
\textcolor{black}{A medical conversation can exceed 30 minutes and surpass 4K in context length. Therefore, we used}
\textcolor{black}{8K context length by applying positional interpolation \cite{chen2023extending} to the base model.} We set the weight decay at 0.1 and a warm-up step count to 50.

\noindent\textbf{Robust Training.} To be tolerant of machine and experiment related mishaps, we used fixed seed, checkpoints, and implemented phased training where we divided the training data into $n$ subsets. If the loss of a particular validation subset started to stabilize, we reduced the sampling rate in the next phase for efficiency.


\noindent\textbf{Data Packing \& Dedup.} We packed data by sentence to fit into max sequence length. We also deduplicated our data to improve data quality \citep{lee2021deduplicating}.



\noindent\textbf{Loss.} For the general corpus including C4, public medical materials, we calculated the gradient on every token. However, on proprietary instruction data, the loss was only calculated on response tokens like \cite{orca}. 

\section{Evaluations}
{This section shows some of our continued pretraining results and evaluation methodology.}

\subsection{Pretraining}

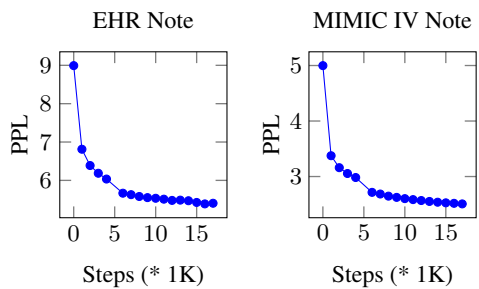
\begin{figure}[t]
     \centering
     \begin{subfigure}[b]{0.2\textwidth}
         \centering
        \begin{tikzpicture}
        \begin{axis}[
            title={EHR Note},
            xlabel={Steps (* 1K)},
            ylabel={PPL},
            width=2.2cm,
            height=2.2cm,
            label style={font=\small},      
            title style={font=\small},      
            inner sep=1.5pt,                  
            outer sep=1.5pt,                  
            scale only axis,                
            ylabel near ticks,
            tick label style={font=\small}, 
            tick style={line width=0.5pt},                 
            label style={font=\small},
        ]
        \addplot [color=blue, mark=*, mark size=1.5pt]
	coordinates { 
(0, 8.99213600158691) (1, 6.81216859817505) (2, 6.38708782196045) (3, 6.18571138381958) (4, 6.03472566604614) (6, 5.66576051712036) (7, 5.62770557403564) (8, 5.57988166809082) (9, 5.55047559738159) (10, 5.53340673446655) (11, 5.50995683670044) (12, 5.4778413772583) (13, 5.48491716384888) (14, 5.4693808555603) (15, 5.42473936080933) (16, 5.38687896728516) (17, 5.40365219116211)

};
        \end{axis}
        \end{tikzpicture}
         \label{fig:y equals x}
     \end{subfigure}
     \begin{subfigure}[b]{0.2\textwidth}
         \centering
        \begin{tikzpicture}
        \begin{axis}[
            title={MIMIC IV Note},
            xlabel={Steps (* 1K)},
            ylabel={PPL},
            width=2.2cm,
            height=2.2cm,
            label style={font=\small},      
            title style={font=\small},      
            inner sep=1.5pt,                  
            outer sep=1.5pt,                  
            scale only axis,                
            ylabel near ticks,
            tick label style={font=\small, inner sep=1pt}, 
            tick style={line width=0.5pt},                 
            label style={font=\small},
        ]

        \addplot  [color=blue, mark=*, mark size=1.5pt]
	coordinates {(0.0, 4.998110) (1.0, 3.375171) (2.0, 3.159490) (3.0, 3.054542) (4.0, 2.982241) (6.0, 2.714931) (7.0, 2.684741) (8.0, 2.646686) (9.0, 2.624459) (10.0, 2.601890) (11.0, 2.584396) (12.0, 2.567634) (13.0, 2.551206) (14.0, 2.537253) (15.0, 2.526099) (16.0, 2.517340) (17.0, 2.506986)};

        \end{axis}
        \end{tikzpicture}
         \label{fig:three sin x}
     \end{subfigure}
        \caption{Pretraining validation perplexity.}
        \vspace{-8pt}
        \label{fig:validation_loss}
\end{figure}

We employed two evaluation methods to monitor pertaining.
\textcolor{black}{Firstly, we measured the perplexity across all the data sources. We used a validation set to track how efficiently the model learns from each source.} Figure \ref{fig:validation_loss} is a subset of evaluations on EHR and MIMIC IV Note. EHR Note is 1K notes sampled from our proprietary dataset, which is the doctors' written notes from real clinic visits. MIMIC IV Note is 1K sampled deidentified critical care notes from the public dataset \cite{johnson2020mimic}. \textcolor{black}{The Figure \ref{fig:validation_loss} shows that as the training continues, the model progressively increases its understanding of both data sets. However, MIMIC IV has a much lower perplexity suggesting that the base LLaMA2 model might have been trained on this dataset during the initial pertaining process.}

Secondly, for a holistic understanding of the generation quality, we used several few-shot \textcolor{black}{(3-shot)} generative tasks for validation, that included:

\noindent \textbf{1) Long text generation}: \textcolor{black}{This task is associated with summarizing different categories of the subjective section of SOAP notes from medical transcripts between doctor and patient. For example:}

\begin{quote}
\textbf{Prompt} Summarize the patient's \textit{chief complaint} from the given text. \newline {Transcript}: \textit{<transcript>} \newline
\textbf{Output} \textit{<response>}
\end{quote}


\noindent \textbf{2) Medium text generation}: This is a question answering task on medical transcript. 
We curated this data by modifying the Alpaca~\citep{alpaca2023stanford} pipeline on the \textcolor{black}{collected transcription} dataset. We queried GPT-4 to generate questions prompting responses ranging from a few words to a full sentence based on the transcription. 
For example:

\begin{quote}
\textbf{Prompt} Identify the patient's current medication.\newline {Transcript}: \textit{<transcript>} \newline
\textbf{Output} \textit{<response>}
\end{quote}
\vspace{-3pt}
\noindent \textbf{3) Short text generation}: This comprises of ROS (Review of System) - related \textcolor{black}{classification} tasks, including questions about body system identification (multi-choice), and absence or presence of symptoms (single-choice). For example:


\begin{quote}
\textbf{Prompt} Is the patient showing signs of depression, like persistent sadness, lack of interest, or appetite changes?
\newline {Transcript}: \textit{<transcript>} \newline
\textbf{Output} \textit{<response>}
\end{quote}
\vspace{-3pt}




We measured Rouge-cls for tasks 1, 2 and accuracy for task 3, to monitor pretraining performance. \textcolor{black}{Each of evaluation dataset has 1000 examples.} 

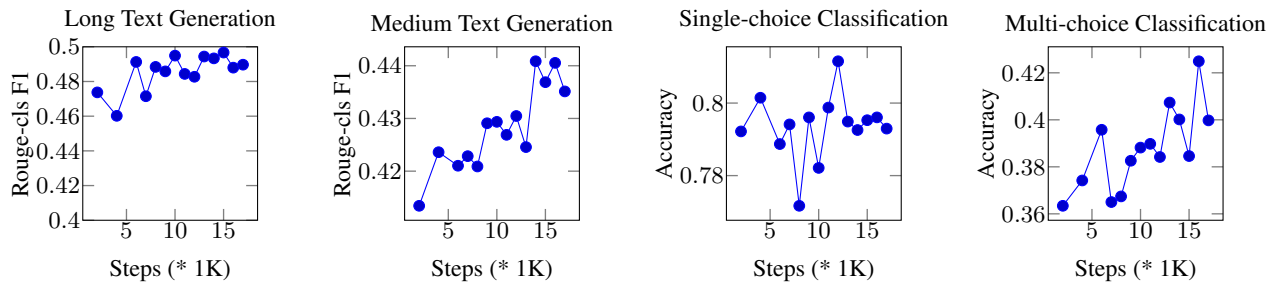
\begin{figure*}[h]
     \centering
     \begin{subfigure}[b]{0.2\textwidth}
         \centering
        \begin{tikzpicture}
        \begin{axis}[
            title={Long Text Generation},
            xlabel={Steps (* 1K)},
            ylabel={Rouge-cls F1},
            width=2.3cm,
            height=2.3cm,
            ymin=0.4, ymax=0.5,
            label style={font=\small},      
            title style={font=\small},      
            inner sep=0pt,                  
            outer sep=0pt,                  
            scale only axis,                
            ylabel near ticks,
            tick label style={font=\small, inner sep=1pt}, 
            tick style={line width=0.5pt},                 
            label style={font=\small},
        ]
        \addplot 
	coordinates {  (2, 0.473723602772819) (4, 0.460201302454042) (6, 0.491272844730344) (7, 0.471496793543925) (8, 0.488409280670971) (9, 0.485826467862144) (10, 0.494910600998884) (11, 0.484365312452636) (12, 0.482716934640849) (13, 0.494385698787548) (14, 0.493327741179185) (15, 0.496694089279637) (16, 0.487973235955864) (17, 0.489682124447526)

};
        \end{axis}
        \end{tikzpicture}
         \label{fig:y equals x}
     \end{subfigure}
     \hfill
     \begin{subfigure}[b]{0.2\textwidth}
         \centering
        \begin{tikzpicture}
        \begin{axis}[
            title={Medium Text Generation},
            xlabel={Steps (* 1K)},
            ylabel={Rouge-cls F1},
            width=2.3cm,
            height=2.3cm,
            label style={font=\small},      
            title style={font=\small},      
            inner sep=0pt,                  
            outer sep=0pt,                  
            scale only axis,                
            ylabel near ticks,
            tick label style={font=\small, inner sep=1pt}, 
            tick style={line width=0.5pt},                 
            label style={font=\small},
        ]

        \addplot 
	coordinates { (2, 0.41342603467900400) (4, 0.4235975526900660) (6, 0.42102730132927400) (7, 0.42284674003345800) (8, 0.4208878180857820) (9, 0.42908990877360) (10, 0.42935689841020200) (11, 0.42689159390800300) (12, 0.43046537796386600) (13, 0.4245685022104740) (14, 0.4408549202949370) (15, 0.4368838894135150) (16, 0.4405463203804250) (17, 0.4351289522967760)};

        \end{axis}
        \end{tikzpicture}
         \label{fig:three sin x}
     \end{subfigure}
     \hfill
     \begin{subfigure}[b]{0.2\textwidth}
         \centering
        \begin{tikzpicture}
        \begin{axis}[
            title={Single-choice \textcolor{black}{Classification}},
            ylabel={Accuracy},
            xlabel={Steps (* 1K)},
            width=2.3cm,
            height=2.3cm,
            label style={font=\small},      
            title style={font=\small},      
            inner sep=0pt,                  
            outer sep=0pt,                  
            scale only axis,                
            ylabel near ticks,
            tick label style={font=\small, inner sep=1pt}, 
            tick style={line width=0.5pt},                 
            label style={font=\small},
        ]
        \addplot
        coordinates {(2, 0.7921826625387) (4, 0.8014705882352940) (6, 0.7886996904024770) (7, 0.7941176470588240) (8, 0.771671826625387) (9, 0.7960526315789470) (10, 0.7821207430340560) (11, 0.7987616099071210) (12, 0.8115325077399380) (13, 0.794891640866873) (14, 0.7925696594427250) (15, 0.7952786377708980) (16, 0.7960526315789470) (17, 0.7929566563467490)};
        \end{axis}
        \end{tikzpicture}
         \label{fig:five over x}
     \end{subfigure}
     \hfill
     \begin{subfigure}[b]{0.2\textwidth}
         \centering
        \begin{tikzpicture}
        \begin{axis}[
            title={Multi-choice \textcolor{black}{Classification}},
            ylabel={Accuracy},
            xlabel={Steps (* 1K)},
            width=2.3cm,
            height=2.3cm,
            label style={font=\small},      
            title style={font=\small},      
            inner sep=0pt,                  
            outer sep=0pt,                  
            scale only axis,                
            ylabel near ticks,
            tick label style={font=\small, inner sep=1pt}, 
            tick style={line width=0.5pt},                 
            label style={font=\small},
        ]
        \addplot
        coordinates {(2, 0.3634185303514380) (4, 0.3742012779552720) (6, 0.3957667731629390) (7, 0.36501597444089500) (8, 0.36741214057508) (9, 0.38258785942492) (10, 0.38817891373801900) (11, 0.38977635782747600) (12, 0.384185303514377) (13, 0.4073482428115020) (14, 0.4001597444089460) (15, 0.3845846645367410) (16, 0.4249201277955270) (17, 0.3997603833865820)};
        \end{axis}
        \end{tikzpicture}
         \label{fig:five over x}
     \end{subfigure}
        \vspace{-10pt}
        \caption{Pretraining validation generation capability monitoring.}
        \vspace{-5pt}
        \label{fig:validation}
\end{figure*}



Figure \ref{fig:validation} demonstrates that our model's performance consistently improved in generating long and medium texts, and in multi-choice classification. However, no significant improvement was observed in single-choice classification. We attribute this to the already high accuracy numbers and the fact that further improvement was noted when the model was separately trained on a smaller related dataset, indicating potential enhancements with scaled-up training.


\subsection{Pretraining Ablation}

\begin{table}
\begin{center} 
\small
\begin{tabular}{c c c c}
\thickhline
 Training & ROS & Long Text & Long Text \\
  data & (multi-choice) & Rouge-1 & Rouge-cls \\
 & (Acc \%)  & (f1 \%) & (f1 \%) \\
 \hline
1B Total  &  \textbf{47.36} & \textbf{44.81} & 41.53\\
MED & 37.85 & 39.44 & 35.91 \\
PUB & 36.81 & \textbf{44.49} & \textbf{42.35} \\
\thickhline
\end{tabular}
\caption {\textbf{Training data ablation results}. The \textbf{MED} dataset is derived from the 1B training dataset by excluding all the public datasets. Similarly, the \textbf{PUB} dataset is produced by removing all medical datasets.}
\label{table:3}
\end{center}
\end{table}

Table \ref{table:3} shows our examination of the effects of varying data proportions using a 1B token dataset, derived from a scaled-down version of our custom 15B dataset on the 7B LLaMA2 model. The ablation study revealed that removing general datasets from the mix detrimentally impacted the model's generative abilities, resulting in decreased summarization quality. We were also able to conclude that the medical datasets indeed improved the model’s understanding of the medical context.
Consequently, we decided to use equal proportions of these datasets during training to maintain the model's generative abilities while improving its understanding of medical contexts.

\begin{table}
\centering
\small
\begin{tabular}{c c c c}
\thickhline
Model & \#Incorrect & \#Irrelevant & \#Missed \\
\hline
Human & 1.20 & \textbf{0} & 11.20 \\
GPT-4 & \textbf{0.80}	& 0.20 & 6.75 \\
Med-PaLM 2 &  1.36 & \textbf{0} &  10.50 \\
GPT-3.5 & 2.00	& 1.71 & 8.50 \\
\dag LLaMA2-chat-13B  & 4.14	& 4.71 & 11.21 \\
\dag PMC-LLaMA-13B & 1.57	& 0.43 & 15.14 \\
\textit{*}LLaMA2-13B  & 1.50	& 0.14 & 9.86 \\
\textit{*}MedLLaMA-13B  & 2.07	& 0.71 & 11.57 \\
\textit{*}Meditron-7B  & 3.00	& 0.57 & 10.64 \\
\hline
\textbf{HEAL} & \textbf{0.85} & \textbf{0.30}	& \textbf{4.30} \\
\thickhline
\end{tabular}
\caption {\textbf{Average entity errors comparison.} Both \textit{*}~and \dag~are fine-tuned models. *~indicates a pretrained model was used as the base, \dag~denotes a fine-tuned instruction model was used as the base.}
\vspace{-5pt}
\label{table:summarization_evaluation}
\end{table}


\subsection{Medical Note Generation}


\noindent\textbf{Evaluation Dataset and Setup.} We compared the HEAL model to several general and medical SOTA models, including the high-end GPT-4, GPT-3.5, and Med-PaLM 2~\citep{singhal2023towards} and other similarly sized open-source medical LLMs, as shown in Table~\ref{table:summarization_evaluation}. We meticulously fine-tuned LLaMA2-Chat-13B~\citep{Touvron2023Llama2O} and the PMC-LLaMA-13B~\citep{wu2023pmc} on medical generative tasks of varying lengths, detailed in Section 3.1 using 10K instruction samples. Pretrained models like LLaMA2-13B~\citep{Touvron2023Llama2O}, MedLLaMA (base model of PMC-LLaMA), and Meditron-7B~\citep{chen2023meditron} were explanation-tuned on our proprietary dataset of 500K examples to enhance their instruction-following capabilities. We also compared these models to human scribes from our production system (medical students who underwent internal scribe training and received monetary compensation for their services). All the models and scribes were evaluated on generating the Subjective and Plan sections of the SOAP medical note using 10 doctor-patient dialogue-style conversations averaging 12 minutes each.

\noindent\textbf{Evaluation Metric.} We leveraged human medical experts to evaluate these models. They developed a rubric note for each transcript, highlighting all essential medical information as separate medical entities. Every entity symbolized a significant sentence or phrase that a healthcare provider needed to approve the note. On average, our experts identified 35 medical entities per transcript. We evaluated the generated notes on three key parameters: \textit{\textbf{Completeness}}, \textit{\textbf{Correctness}}, and \textit{\textbf{Conciseness}} as outlined in \citep{clinical-text-summ} using the following metrics: 

\noindent 1) \textit{\textbf{Missed Information}} refers to the entities omitted in the test note relative to the rubric note. This metric reflects the test note's completeness.

\noindent 2) \textit{\textbf{Incorrect Information}} implies the entities inaccurately captured by the test note. This metric is critical in healthcare where information accuracy is essential, as misinformation can erode trust in AI. 

\noindent 3) \textit{\textbf{Irrelevant information}} refers to extraneous elements in the test note not linked to the rubric note. As lengthy medical notes require more time for review, it's crucial to reduce irrelevant information.

\noindent \textbf{Results and Analysis.} Table~\ref{table:summarization_evaluation} compares the performance of our HEAL model, other models, and human scribes. Notably, HEAL surpasses all other models in the Missed Information metric, indicating a superior ability to identify and summarize critical medical information. We attribute this improved performance to our continued pretraining approach using complex medical instructions. We also observed some inaccuracies due to ASR (Automatic Speech Recognition) errors, yet both our model and GPT-4 excelled at correcting these mistakes. Human scribes and Med-PaLM 2 created concise notes but missed vital medical details. Other models, such as GPT 3.5, MedLLaMa, and LLaMA2-chat, struggled to grasp real-world conversation nuances, as shown by their high Incorrect and Missed Information scores. Overall, our model shows exceptional performance in all metrics of the task, outperforming both human scribes and other fine-tuned models.

In our detailed quality evaluation, we found that a human scribe takes about 1.67 times longer than the audio recording to create a medical note. However, AI models can generate the same note almost instantly, demonstrating the efficiency and time-saving capabilities of AI in medical transcription.

\subsection{Public Benchmark}
Although HEAL is specifically designed for medical note summarization, we also tested its performance against other LLMs on two popular medical benchmarks to evaluate its efficiency in other medical tasks.

\noindent \textit{PubMedQA~\citep{pubmedqa}} A biomedical QA task 
to answer research questions with yes/no/maybe using the corresponding PubMed paper snippets.

\noindent \textit{MedQA~\citep{medqa}} Multi-choice questions extracted from US Medical License Exams.

{\tiny
\begin{table}[t]
\centering
\scriptsize
\begin{tabular}{c c c c c c}
\thickhline
Dataset & LLaMA2 & PMC- & GPT-4 & Med- & HEAL \\
& 13B & LLaMA & (5-shot) & PaLM 2 & 13B \\
&  & 13B &  & (best) & \\
\hline
PubMedQA & 76.40 & 77.90 & 75.2 & \textbf{81.8} & 78.4 \\
MedQA & 45.48 & 56.36 & 81.4 & \textbf{86.5} & 47.2 \\
\thickhline
\end{tabular}
\caption {Accuracy (\%) on PubMedQA and MedQA.}
\vspace{-5pt}
\label{table:public_benchmark}
\end{table}
}

In PubMedQA, Med-PaLM 2 with the best prompting strategy \citep{singhal2023towards} took advantage of its huge size and further tuning on PubMedQA data to achieve the highest score. As shown in Table ~\ref{table:public_benchmark}, HEAL achieved 78.4\% accuracy after tuning, which surpasses GPT-4's performance~\citep{nori2023capabilities}, fine-tuned LLaMA2 and even PMC-LLaMA ~\citep{wu2023pmc} which is further tuned on 75B PubMed data. Our improved performance can be attributed to our proprietary medical instruction data on conversational data which focuses more on medical understanding. 




In MedQA, we attained a 47.2\% accuracy rate, surpassing the LLaMA2 13B model yet falling short of PMC-LLaMA. MedQA focuses on medical reasoning, requiring the model to recall medical knowledge and derive diagnoses or solutions from specified problems. Larger models like GPT-4, Med-PaLM 2, or those trained with vast amounts of data hold an inherent advantage in this task. HEAL, which is geared towards interpreting medical conversations, does not align with this task, yielding suboptimal performance on this dataset.




\section{Conclusion}


This paper presents our work of developing a medical LLM capable of comprehending and summarizing medical conversation. As a result, this is the first model, with significantly fewer parameters, to outperform 
\textcolor{black}{humans, existing medical LLMs including Med-PaLM 2, PMC-LLaMA and perform on par with GPT-4.}
\textcolor{black}{Our evaluation shows} that even small-scale continued pretraining of smaller LLMs can show impressive gains. 
\textcolor{black}{We believe that scaling up our training can further improve results. Our work presents a promising development in healthcare documentation and other medical areas.}


\vspace{-3pt}
\section{Related Work}

\noindent\textbf{Medical LLMs.} 
Various medical LLMs such as 
MedGPT~\citep{kraljevic2021medgpt}, 
and Med-PaLM 2~\citep{singhal2023towards} show how training on various medical datasets, improves model's performance on medical knowledge understanding tasks.
MEDITRON-70B~\citep{chen2023meditron}, the state-of-the-art open-source LLM  
and PMC-LLaMA~\citep{wu2023pmc}
demonstrates the 
effectiveness of task-specific fine-tuning and instruction tuning.

\noindent\textbf{Domain adaption LLM.} 
As demonstrated by \citep{dontstop}, \citep{scibert}, continued pretraining on unlabeled, domain-specific data boosts model performance on domain tasks, providing a practical solution when resources for scratch domain-adaptive pretraining are limited.

\noindent\textbf{Medical Note Generation.} Prior work by \citep{leveraging-pretrained}, \citep{clinical-text-summ} 
demonstrated the feasibility of using Language Models to generate medical summaries from dialogues. However, they primarily aimed at producing partial notes or semi-automated methods requiring human involvement, rather than comprehensive, provider-ready reports.

\noindent\textbf{Explanation tuning.}
Orca~\citep{orca, orca2} models showcased that smaller Language Models capable of sound reasoning can efficiently perform complex tasks. They were trained by explanation tuning a LLaMA2 13B model~\citep{touvron2023llama} using bigger models like GPT4 as a teacher. 




\section{Ethical Considerations}
\textcolor{black}{All the data processing and experiments were done in HIPAA-compliant environment. We deidentified clinical data to remove any PHI information as per our data compliance agreement. HEAL is only used for internal medical tasks like summarization, transcription based Q\&A, and note review. All prompts are audited to prevent unintentional usage.}

\section{Limitations}
\textcolor{black}{Our design focuses on contextual comprehension and summarization of transcripts, and can be further improved on MedQA or similar benchmarks with training on more medical data.}
Future projects could explore utilizing more sophisticated base models, curating higher quality data with a balanced mix of medical knowledge and reasoning content, and scaling up the experiment.





\bibliography{custom, others}

\end{document}